\documentclass[letterpaper]{article} 
\usepackage{aaai2026}  
\usepackage{times}  
\usepackage{helvet}  
\usepackage{courier}  
\usepackage[hyphens]{url}  
\usepackage{graphicx} 
\usepackage{natbib}  
\usepackage{caption} 
\usepackage{algorithm}
\usepackage{algorithmic}
\usepackage{courier}
\usepackage{amsmath}
\usepackage{bm}
\usepackage{tikz}
\usepackage{xcolor}
\usepackage{booktabs}
\usepackage{multirow}
\usepackage{amsfonts}

\definecolor{first_Blue}{RGB}{72,116,203}
\definecolor{first_Orange}{RGB}{238,130,47}
\definecolor{first_Yellow}{RGB}{242,186,2}
\definecolor{first_Green}{RGB}{117,189,66}

\usepackage{newfloat}
\usepackage{listings}
\DeclareCaptionStyle{ruled}{labelfont=normalfont,labelsep=colon,strut=off} 
\floatstyle{ruled}
\newfloat{listing}{tb}{lst}{}
\floatname{listing}{Listing}
\title{Pansharpening for Thin-Cloud Contaminated Remote Sensing Images: A Unified Framework and Benchmark Dataset}
\author{
    Songcheng Du\textsuperscript{\rm 1}\equalcontrib, Yang Zou\textsuperscript{\rm 1}\equalcontrib, Jiaxin Li\textsuperscript{\rm 2}, Mingxuan Liu\textsuperscript{\rm 1}, Ying Li\textsuperscript{\rm 1}\thanks{Corresponding author.}, Changjing Shang\textsuperscript{\rm 3}, Qiang Shen\textsuperscript{\rm 3}
    }
\affiliations{
    \textsuperscript{\rm 1}Northwest Polytechnical University\\
    \textsuperscript{\rm 2}Chongqing University of Posts and Telecommunications\\
    \textsuperscript{\rm 3}Aberystwyth University\\
    dusongcheng@mail.nwpu.edu.cn, archerv2@mail.nwpu.edu.cn, lijx@cqupt.edu.cn, 2023302820@mail.nwpu.edu.cn, lybyp@nwpu.edu.cn, cns@aber.ac.uk, qqs@aber.ac.uk
}

\usepackage{bibentry}

\begin{document}

\maketitle
\begin{abstract}
Pansharpening under thin cloudy conditions is a practically significant yet rarely addressed task, challenged by simultaneous spatial resolution degradation and cloud-induced spectral distortions. Existing methods often address cloud removal and pansharpening sequentially, leading to cumulative errors and suboptimal performance due to the lack of joint degradation modeling. To address these challenges, we propose a Unified Pansharpening Model with Thin Cloud Removal (Pan-TCR), an end-to-end framework that integrates physical priors. Motivated by theoretical analysis in the frequency domain, we design a frequency-decoupled restoration (FDR) block that disentangles the restoration of multispectral image (MSI) features into amplitude and phase components, each guided by complementary degradation-robust prompts: the near-infrared (NIR) band amplitude for cloud-resilient restoration, and the panchromatic (PAN) phase for high-resolution structural enhancement. To ensure coherence between the two components, we further introduce an interactive inter-frequency consistency (IFC) module, enabling cross-modal refinement that enforces consistency and robustness across frequency cues.  Furthermore, we introduce the first real-world thin-cloud contaminated pansharpening dataset (PanTCR-GF2), comprising paired clean and cloudy PAN-MSI images, to enable robust benchmarking under realistic conditions. Extensive experiments on real-world and synthetic datasets demonstrate the superiority and robustness of Pan-TCR, establishing a new benchmark for pansharpening under realistic atmospheric degradations. 
\end{abstract}

\begin{links}
\link{Code}{https://github.com/dusongcheng/PanTCR-GF2}
\end{links}

\begin{figure}
    \centering
    \includegraphics[width=1\linewidth]{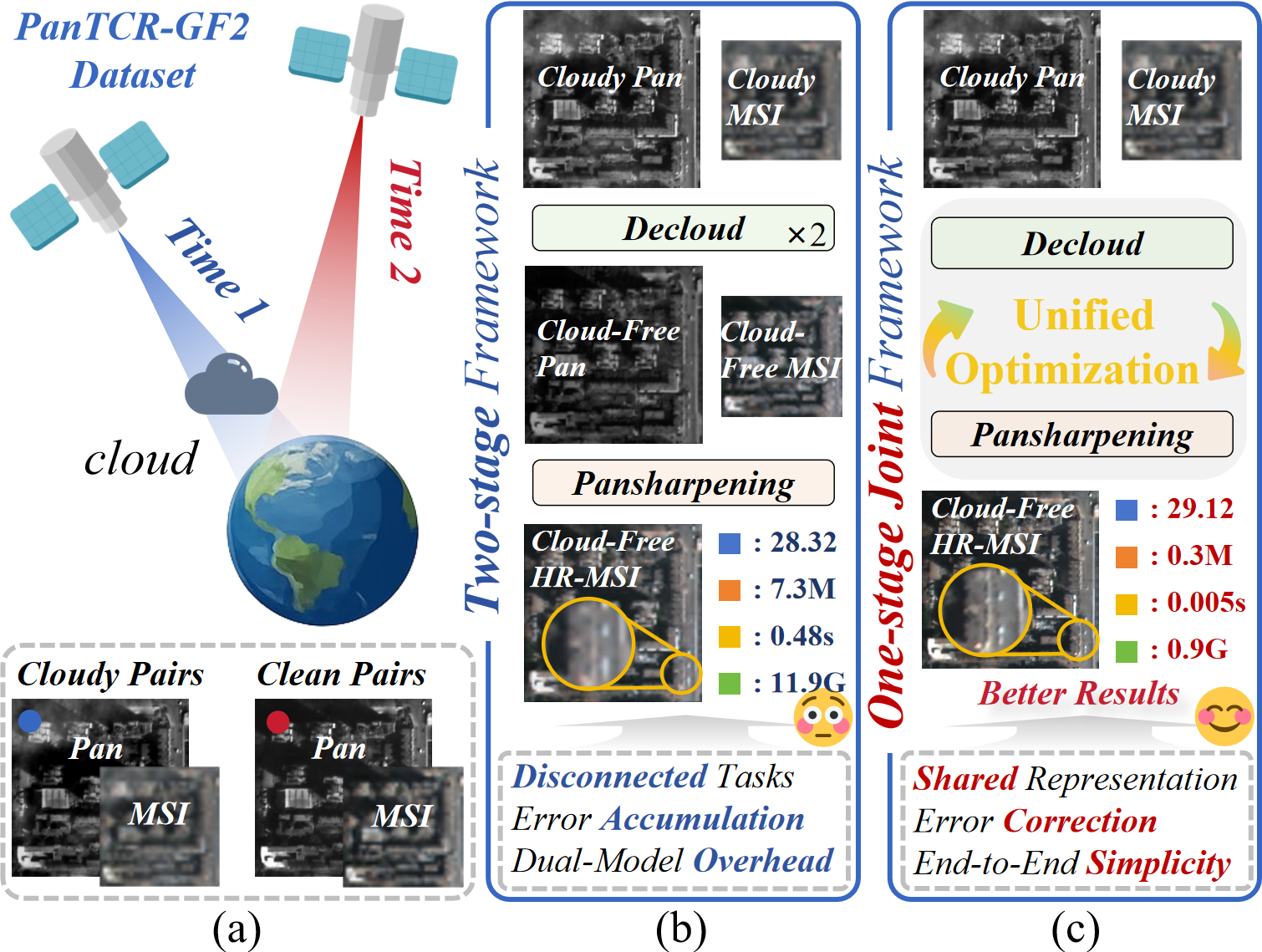}
    \caption{Collection of the first thin-cloud pansharpening dataset (a) and comparison of processing paradigms and performance between existing methods (b) and proposed approach (c). 
    \({\color{first_Blue}\rule{0.2cm}{0.2cm}}\): PSNR, 
    \({\color{first_Orange}\rule{0.2cm}{0.2cm}}\): Parameters, 
    \({\color{first_Yellow}\rule{0.2cm}{0.2cm}}\): Test Time, 
    \({\color{first_Green}\rule{0.2cm}{0.2cm}}\): FLOPs.}
    \label{fig_first}
\end{figure}

\section{Introduction}
Pansharpening aims to generate high-resolution multispectral images (MSIs) by fusing the spectral information of low-resolution MSIs with the spatial details of panchromatic (PAN) images. This technique is crucial for a wide range of applications~\cite{li2025exploring2,li2025stitchfusion,lan2025acam}, including land use monitoring, precision agriculture, and environmental surveillance~\cite{xi2025hypertafor,peng2025deep,du2023degradation,li2022hasic}. However, real-world satellite images are frequently contaminated by cloud cover, which not only degrades the spatial textures but also distorts spectral reflectance in a non-uniform manner.

Despite the widespread occurrence of cloud, particularly thin-cloud, in optical satellite imagery, most existing pan-sharpening methods assume clear-sky conditions~\cite{wfanet,arconv,du2024unsupervised,li2025maris,zou2024contourlet,li2026CIGPose,lan2026reco}, making them unreliable under cloud-contaminated scenarios. A common workaround is to perform cloud removal and pansharpening sequentially, where cloud-free MSI and PAN are first reconstructed using two separate cloud removal networks, followed by conventional pansharpening~\cite{li2017improvement,meng2018pansharpening,zhang2026igasa,wang2024low}. As illustrated in Fig.~\ref{fig_first}(b), to implement such an approach, a state-of-the-art (SOTA) decloud algorithm~\cite{thiefcloud} and a leading pansharpening method~\cite{wfanet} may be employed. While intuitive, this two-stage design suffers from several drawbacks: 1) \textbf{Cascaded error accumulation.} Artifacts arising from the cloud removal stage, including residual clouds and spectral inconsistencies, may be propagated and further amplified during pansharpening, degrading the final result. 2) \textbf{High computational overhead.} The cascade of two separate networks significantly raises inference time and memory consumption, making deployment impractical in real-time or resource-constrained environments. 

Besides, following the cascade approach, thin-cloud removal and pansharpening are typically conducted by separate models trained under different settings and objectives, resulting in inconsistent intermediate representations and suboptimal final outputs. Although several recent works have proposed joint learning strategies for other multimodal fusion tasks~\cite{feijoo2025darkir,su2025unifuse,li2023text,liu2024promptfusion,gao25,lan2026performance}, few studies have explicitly addressed the unified modeling of thin-cloud and resolution degradations within a single pansharpening framework. This lack of research attention limits the robustness of existing methods in practical scenarios where thin-cloud contamination is the norm rather than the exception.

Therefore, designing an end-to-end framework that simultaneously addresses both resolution degradation and thin-cloud interference is essential. However, achieving this requires: (1) effectively leveraging complementary cues from heterogeneous inputs (e.g., PAN, near-infrared image (NIR)), and (2) disentangling the above two degradations in a principled manner. To this end, we propose \textbf{Pan-TCR}, the first unified pansharpening framework (see Fig.~\ref{fig_first}(c)) that explicitly models both degradations within a frequency-prompt guided formulation, offering a principled solution tailored to real-world satellite imagery.

\begin{figure*}
    \centering
    \includegraphics[width=0.96\linewidth]{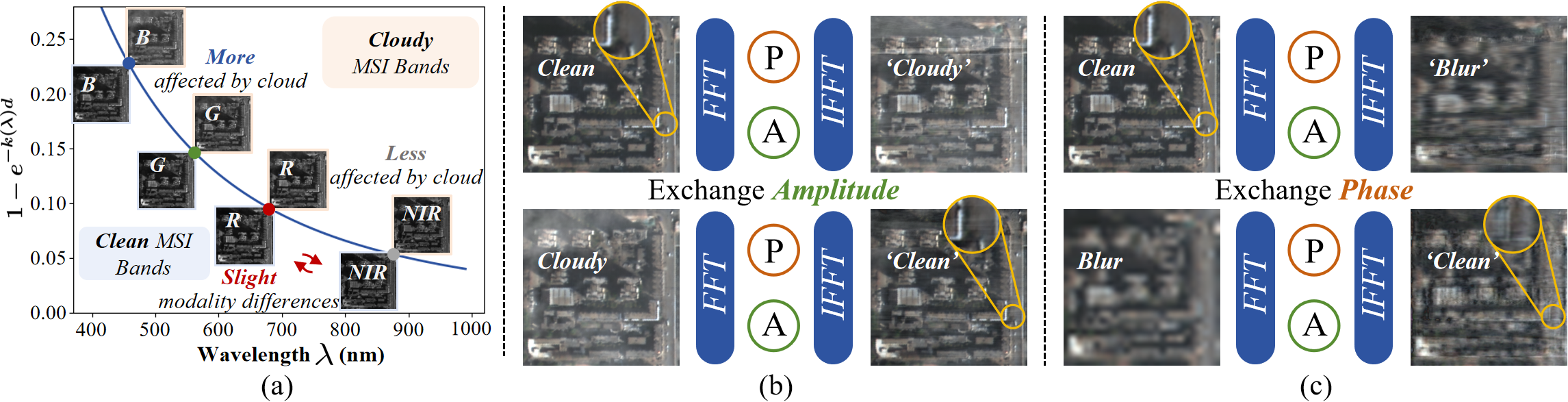}
    \caption{Motivations. (a) Spectral sensitivity to thin-cloud across different wavelengths \(\lambda\), modeled via atmospheric scattering model \(E(\lambda)\), where \(E_0(\lambda)\), \(L_\infty(\lambda)\) and \(e^{-k(\lambda) d}\) denote clean surface radiance, airlight and transmittance, respectively, and \(1-e^{-k(\lambda) d}\) denotes cloud interference. \textbf{NIR band is less affected by cloud scattering} than shorter wavelengths (e.g., B, G, R). Although NIR is cloud-robust, its \textbf{modality difference} from visible bands presents challenges for spectral consistency. (b) \textbf{The amplitude component} primarily reflects global intensity distributions and \textbf{is highly sensitive to cloud-induced degradation}, whereas (c) \textbf{the phase component} encodes fine structural details and \textbf{is more affected by resolution degradation}. Moreover, directly exchanging amplitude or phase between images can result in \textbf{inter-frequency inconsistency artifacts} (highlighted by yellow circles).}
    \label{fig_prior}
\end{figure*}

Our key insight stems from the observation that the NIR band within MSI is inherently more robust to thin-cloud interference (Fig.~\ref{fig_prior} (a)), while the PAN image offers high-resolution structures yet is vulnerable to thin-cloud interference similar to visible bands. Motivated by their distinct frequency-domain characteristics (Fig.~\ref{fig_prior}(b) and (c)), we treat the NIR amplitude and PAN phase as two degradation-robust frequency prompts, respectively guiding cloud-resilient amplitude restoration and structure-preserving phase enhancement. 

To implement this design, we propose a \textbf{Frequency-Decoupled Restoration (FDR) Block} that performs component-wise restoration on amplitude and phase, enabling simultaneous suppression of cloud artifacts and recovery of structural details. To enhance consistency between these frequency components (highlighted in Fig.~\ref{fig_prior}(b) and (c)), we embed an \textbf{Interactive inter-Frequency Consistency (IFC) Module} within the FDR block to enhance consistency between these components through interactive cross-modal refinement. Additionally, we introduce a \textbf{Modality-Adaptive Frequency Gating (MAFG) Module} to dynamically suppress modality-induced spectral outliers, enhancing cross-modal consistency between NIR and visible bands. To further refine spectral representations beyond the frequency domain~\cite{li2025frequency}, a \textbf{Spectral Enhancement (SE) Module} is incorporated to provide fine-grained spectral-wise correction complementary to frequency-based restoration.

Furthermore, to support the current development in this area, we construct the first real-world dataset for pansharpening under thin-cloud conditions, named \textbf{PanTCR-GF2} (Fig.~\ref{fig_first}(a)). It contains 15,603 high-quality, aligned pairs of cloudy/clean PAN and MSI images across five representative land cover categories, supporting comprehensive benchmarking under realistic atmospheric interference.

Our contributions made in this work are as follows:

\begin{itemize}
\item We propose \textbf{Pan-TCR}, the first unified end-to-end framework that jointly addresses thin-cloud removal and pansharpening within a frequency-prompt restoration formulation tailored for real-world remote sensing imagery.

\item We introduce the \textbf{Frequency-Decoupled Restoration Block}, which leverages NIR amplitude and PAN phase as degradation-robust frequency prompts to perform targeted restoration on amplitude and phase components.

\item We build \textbf{PanTCR-GF2}, the first real-world benchmark dataset for thin-cloud contaminated pansharpening, enabling systematic evaluation of fusion methods under atmospheric degradation.

\item Our method achieves state-of-the-art performance across multiple benchmarks while being more efficient in terms of model size, FLOPs, and inference time.

\end{itemize}

\section{Related Work}
\subsection{Pan-sharpening}

Early pansharpening techniques, such as component substitution~\cite{carper1990use}, multi-resolution analysis~\cite{aiazzi2006mtf}, and variational optimization~\cite{ballester2006variational}, are prone to spectral distortions due to their limited capacity for semantic modeling. With the advent of deep learning~\cite{tcsvt1,gao,gao2025knowledge,li2025exploring1,shen2025fine,li2026toward}, CNN-based methods~\cite{li2026comprehensive,li2025mmt,li2025srkd} like PNN~\cite{masi2016pansharpening}, PanNet~\cite{yang2017pannet}, and BDPN~\cite{zhang2019pan} marked a turning point, achieving significant fusion improvements. Later works introduced adaptive convolutions (e.g., LAGConv~\cite{jin2022lagconv}, CANNet~\cite{cannet}) and frequency-domain designs (e.g., SFINet~\cite{sfinet}, BiMPan~\cite{bimpan}) to further enhance spatial fidelity. FusionMamba~\cite{fusionmamba}  and PanMamba~\cite{panmamba} attempt to leverage the Mamba architecture for improving efficiency. Recently, diffusion models~\cite{ssdiff,kim2025u,feng2026sr3r,wangefficient} have been explored for their generative capacity. However, nearly all existing methods assume cloud-free conditions~\cite{li2023progressive,du2026unsupervised}, limiting their robustness in real-world degraded scenarios.

\subsection{Thin-Cloud Removal}
Unlike thick clouds that completely obscure the surface, thin clouds permit partial signal transmission, thereby allowing algorithmic restoration.
Traditional methods rely on priors like spectral correlation~\cite{zhang2002image,li2022drcr}, dark channel~\cite{he2010single}, or frequency filtering~\cite{shen2014effective}, but struggle with complex cloud patterns. Learning-based approaches offer robust solutions, utilizing residual learning~\cite{li2019thin,feng2024srgs}, cloud prior learning (e.g., ThiefCloud~\cite{thiefcloud}), attention mechanisms~\cite{xu2025tanet}, wavelet and Fourier decompositions (e.g., WaveCNN-CR~\cite{zi2023wavelet}). Recent works also explore Mamba-based architectures (e.g., CR-Famba~\cite{liu2025cr}), and transformer variants~\cite{liu2024cascaded,dai2024tcme,song2023vision,li2025decloudformer} for improved feature extraction and global context modeling. Moreover, generative paradigms such as GANs~\cite{tan2024unsupervised}, VAEs~\cite{ding2022uncertainty}, and diffusion models~\cite{yu2025dc4cr,li2025difiisr}, further enhance realism in reconstruction.

\section{Proposed Method}
\subsection{Framework Overview}
\begin{figure*}
    \centering
    \includegraphics[width=0.96\linewidth]{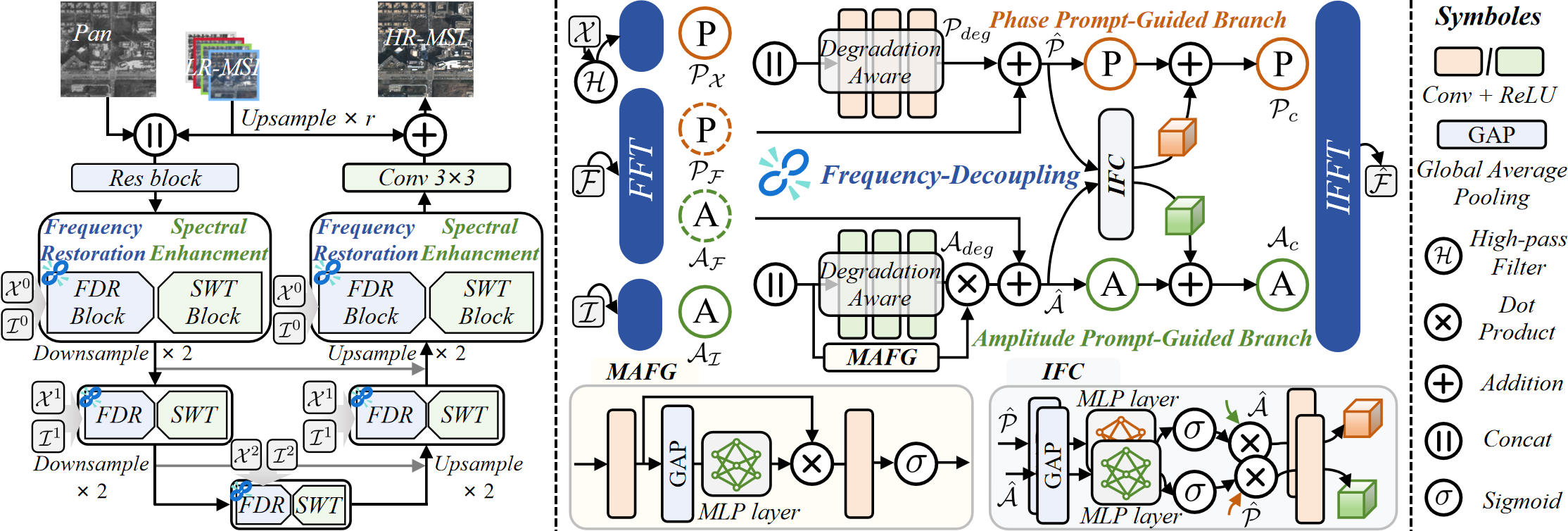}
    \caption{Overview of proposed framework. The left panel illustrates the overall architecture, which takes a stacked input of LR-MSI and PAN images and progressively reconstructs cloud-free high-resolution MSI (HR-MSI) outputs. The central panel illustrates the core frequency-decoupled restoration (FDR) block, which incorporates key sub-modules, including modality-aware frequency gating (MAFG) and interactive inter-frequency consistency (IFC) to guide amplitude and phase restoration via degradation-robust frequency prompts. The right panel provides the symbols used throughout the pipeline.
    }
    \label{fig_network}
\end{figure*}

As illustrated in Fig.~\ref{fig_network}, we propose \textbf{Pan-TCR}, a unified end-to-end framework that simultaneously performs pansharpening and thin-cloud removal by leveraging complementary frequency and spectral restoration strategies. Given a cloud-induced low-resolution multispectral image (LR-MSI, \(\mathcal{Y}\in \mathbb{R}^{\frac{H}{r}\times \frac{W}{r}\times C}\)) and a cloud-induced high-resolution PAN image (\(\mathcal{X}\in \mathbb{R}^{H\times W\times 1}\)), the model generates a high-quality, cloud-free high-resolution MSI (HR-MSI, \(\mathcal{Z}\in \mathbb{R}^{H\times W\times C}\)).

To align spatial dimensions, the LR-MSI is first upsampled using bicubic interpolation and concatenated with the PAN image. The fused input is passed through a shallow residual block to extract initial features \(\mathcal{F}^{0}_E\), which are then processed through a three-stage encoder–decoder pipeline~\cite{wang2025vehiclemae,Che2026LEMON}. Each stage integrates two components: the frequency-decoupled restoration (FDR) block, which leverages PAN (\(\mathcal{X}^i\), \(i \in [0,1,2]\)) and NIR (\(\mathcal{I}^i\)) as two degradation prompts for frequency-domain restoration across different resolutions, and the spectral enhancement (SE) module, which refines spectral consistency.

The FDR block is designed to separately address cloud-induced amplitude degradation and resolution-related phase degradation. Different from previous frequency-based methods (e.g., DarkIR~\cite{feijoo2025darkir}, FourLLIE~\cite{wang2023fourllie}, and FECNet~\cite{huang2022deep}) that process amplitude and phase sequentially, risking error accumulation, FDR block employs a dual-branch structure for parallel processing. The NIR band amplitude serves as a cloud-robust prompt to guide cloud-contaminated restoration, while the PAN phase acts as a resolution-robust prompt for spatial enhancement. Both guidance prompts (\(\mathcal{X}\), \(\mathcal{I}\)) are spatially aligned with the input feature \(\mathcal{F}\). Using Fast Fourier Transform (FFT), we decouple amplitude and phase components for targeted restoration in the frequency domain.

\textbf{Phase Prompt-Guided Branch.} To model resolution degradation while minimizing thin-cloud interference, we first extract high-frequency structures from the PAN image using a contrast-aware learnable high-pass filter (\(\mathcal{H}(\cdot)\)):
\begin{equation}
\mathcal{X}_{\text{filtered}} = \mathcal{X} \otimes \sigma \left( |\mathcal{X} - \text{AvgPool}(\mathcal{X})| \right),
\end{equation}
where \(\sigma\), \(\otimes\), \(\text{AvgPool}(\cdot)\), and \(|\cdot|\) denote the sigmoid~\cite{zhang2026chain}, dot product, average pooling, and absolute operation, respectively. This enhances local structural contrast while suppressing low-frequency noise such as clouds.

Next, we apply FFT to extract phase components \(\mathcal{P}_\mathcal{X}\) and \(\mathcal{P}_\mathcal{F}\) from \(\mathcal{X}_{\text{filtered}}\) and input features \(\mathcal{F}\). The concatenated phase inputs are fed into a lightweight Degradation Aware Module (DAM), comprising three convolutional layers with ReLU activations, to obtain phase residual \(\mathcal{P}_{deg}\), and the enhanced phase representation can be obtained:
\begin{equation}
\hat{\mathcal{P}} = \mathcal{P}_\mathcal{F} + \mathcal{P}_{deg}.
\end{equation}

By using \(\mathcal{P}_\mathcal{X}\) as a prompt, unlike complex architectures in prior works (e.g., DarkIR~\cite{feijoo2025darkir}, FourierISP~\cite{he2024enhancing}, FourLLIE~\cite{wang2023fourllie}), our lightweight design efficiently captures localized phase discrepancies, ensuring accurate spatial restoration with reduced computational complexity.

\textbf{Amplitude Prompt-Guided Branch.} This branch mitigates cloud-induced degradation using the NIR band amplitude as a cloud-robust prompt. We apply FFT to obtain amplitude components from the cloud-robust NIR band \(\mathcal{I}\) and the degraded input feature map \(\mathcal{F}\), denoted as \(\mathcal{A}_\mathcal{I}\) and \(\mathcal{A}_\mathcal{F}\). Concatenated inputs are passed through a DAM residual predictor to estimate the amplitude residual \(\mathcal{A}_{deg}\). 

Due to inherent modality discrepancies between NIR and visible bands, direct residual modeling may introduce spectral artifacts. To mitigate this issue, we propose the modality-adaptive frequency gating (MAFG) module, which performs channel-wise spectral gating to suppress modality-inconsistent features and enhance cross-modal compatibility~\cite{zhang2025ascot,che2026stitch}. Specifically, the concatenated amplitude features \([\mathcal{A}_\mathcal{I}, \mathcal{A}_\mathcal{F}]\) are compressed via a \(1\times1\) convolution, followed by global average pooling to capture spectral statistics. Further, an MLP with sigmoid activation generates a gating vector that reflects the spectral relevance and reliability of each channel in the amplitude domain. This gating vector is subsequently used to modulate \(\mathcal{A}_{\text{deg}}\), adaptively suppressing noisy or inconsistent components while preserving cloud-resilient and spectrally aligned information. The enhanced amplitude representation is given by:
\begin{equation}
    \hat{\mathcal{A}} = \mathcal{A}_{\mathcal{F}} + f_{gate}([\mathcal{A}_\mathcal{I}, \mathcal{A}_\mathcal{F}]) \otimes \mathcal{A}_{deg},
\end{equation}
where \([\cdot,\cdot]\) and \(\otimes\) represent the concatenation and dot product operation, \(f_{gate}(\cdot)\) denotes the MAFG function.

\textbf{Interactive inter-Frequency Consistency (IFC) Module.} 
The FDR block employs a parallel dual-branch strategy to recover amplitude and phase features, guided by distinct degradation-robust priors. However, this independent processing may introduce subtle inconsistencies between the two frequency components, potentially limiting the fidelity of the final reconstruction.

To promote frequency-domain coherence, we embed the IFC module as an interactive refinement unit. It employs a bidirectional cross-modulation mechanism that allows the amplitude and phase branches to guide each other via global feature cues.

Concretely, given the enhanced phase \(\hat{\mathcal{P}}\in \mathbb{R}^{H\times W\times C}\) and amplitude \(\hat{\mathcal{A}}\in \mathbb{R}^{H\times W\times C}\), we first extract global representations using global average pooling, followed by MLP-based projections and sigmoid activation to obtain modulation weights:
\begin{equation}
\small
     \mathbf{w}_{\mathcal{P}} = \sigma(\text{MLP}_{\mathcal{P}}(\text{GAP}(\hat{\mathcal{P}}))), 
     \mathbf{w}_{\mathcal{A}} = \sigma(\text{MLP}_{\mathcal{A}}(\text{GAP}(\hat{\mathcal{A}}))), 
\end{equation}
where \(\mathbf{w}_{\mathcal{P}}\in \mathbb{R}^{1\times 1\times C}\), \(\mathbf{w}_{\mathcal{A}}\in \mathbb{R}^{1\times 1\times C}\) serve as cross-frequency modulation weights that encode amplitude-to-phase and phase-to-amplitude guidance, respectively, and are then used to perform channel-wise modulation on the opposite component to capture inter-frequency residuals:
\begin{equation}
    \mathcal{P}_{res} = \mathbf{w}_{\mathcal{A}} \otimes \hat{\mathcal{P}},
    \mathcal{A}_{res} = \mathbf{w}_{\mathcal{P}} \otimes \hat{\mathcal{A}},
\end{equation}
where \(\otimes\) denotes channel-wise multiplication via broadcasting. Finally, these residuals are added back to their respective original components to yield inter-frequency consistent representations:
\begin{equation}
    \mathcal{P}_{c} = \mathcal{P}_{res}+\hat{\mathcal{P}},
    \mathcal{A}_{c} = \mathcal{A}_{res}+\hat{\mathcal{A}}.
\end{equation}

This cross-frequency interaction harmonizes the amplitude and phase domains, mitigating internal inconsistencies and enhancing robustness. The refined components (\(\mathcal{P}_{c}\),\(\mathcal{A}_{c}\)) are subsequently combined via Inverse Fast Fourier transform (IFFT) to reconstruct a spatially consistent and cloud-free image representation \(\hat{\mathcal{F}}\).

\noindent \textbf{Spectral Enhancement (SE) Module} 

While frequency-domain processing effectively addresses global degradation patterns such as cloud-induced amplitude suppression and resolution-related phase distortions, it often falls short in modeling fine-grained spectral dependencies and semantic inter-band relationships. Although the proposed MAFG module introduces channel-wise gating to mitigate modality discrepancies in the amplitude spectrum, its low-rank and frequency-based nature inherently limits its ability to perform context-aware spectral refinement.

To maintain efficiency while addressing this limitation, we introduce a lightweight SE module in the image domain to refine the spectral fidelity of the restored feature \(\hat{\mathcal{F}}\). Using a simple spectral-wise transformer (SWT)~\cite{swt}, it models long-range dependencies and adaptively recalibrates spectral responses, providing complementary spectral refinement beyond frequency-guided restoration.

\subsection{PanTCR-GF2 Dataset}
\begin{figure}
    \centering
    \includegraphics[width=1\linewidth]{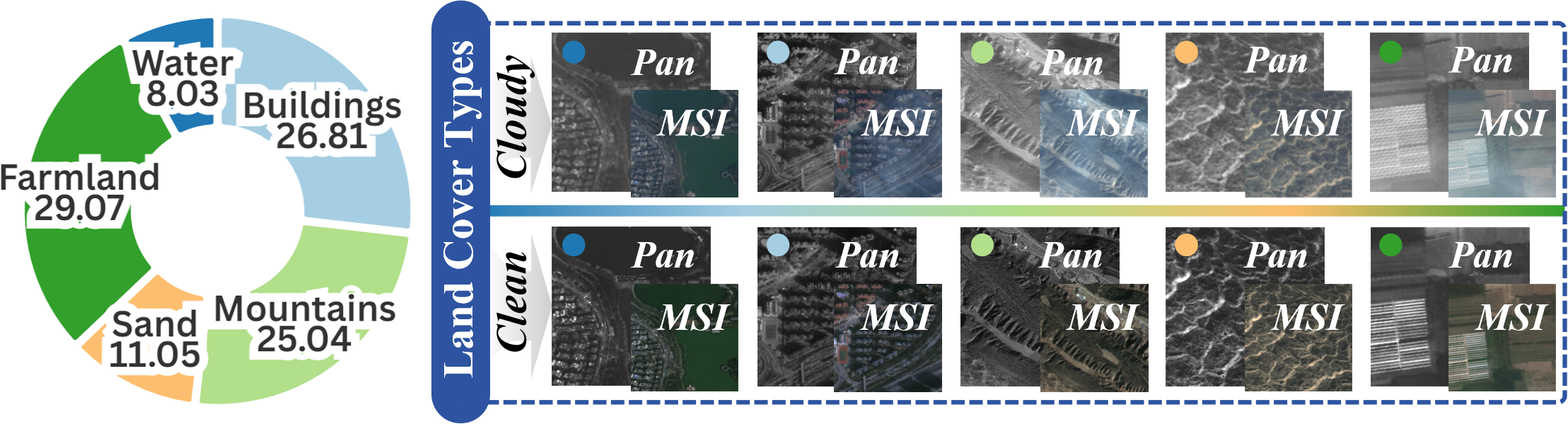}
    \caption{Overview of PanTCR-GF2 dataset: Distribution of land cover types with representative paired examples.
    }
    \label{fig_dataset}
\end{figure}

To support research on pansharpening under realistic atmospheric conditions, we construct the first real-world dataset tailored for this task, named \textbf{PanTCR-GF2}.

\textbf{Data Collection.}
We construct the PanTCR-GF2 dataset using dual-temporal observations captured from the GaoFen-2 satellite, which provides 1-meter PAN and 4-meter LR-MSI images covering four bands: Blue, Green, Red, and NIR. Specifically, we collect 20 scene pairs from identical geographic locations, captured within a one-month interval, one with thin-cloud cover and the other cloud-free, ensuring robust thin-cloud pansharpening evaluation.

\textbf{Data Preparation.}
To construct aligned image pairs, we first extract spatially overlapping regions from the dual-temporal scenes, followed by sub-pixel registration across four images (clean MSI, clean PAN, cloudy MSI, and cloudy PAN). Histogram matching is then applied to mitigate radiometric discrepancies caused by illumination variations or seasonal changes. After registration, the images are partitioned into fixed-size patches, and samples exhibiting significant land surface changes, cloudless or thick clouds are excluded to ensure data quality.

The final dataset contains 15,603 precisely aligned cloudy and clean PAN-MSI pairs, including 14,065 for training, 1,438 for validation, 80 reduced-resolution test patches (128×128), and 20 full-resolution test scenes (512×512), following the PanCollection processes. Fig.\ref{fig_dataset} presents statistical distributions of land cover categories, along with representative examples.
\begin{table*}[]
	\centering
	\resizebox{\linewidth}{!}{\begin{tabular}{cccccccccccc}
		\toprule[1.2pt]
		\multirow{2}{*}{Method} & \multirow{2}{*}{Param} & \multirow{2}{*}{FLOPs} & \multirow{2}{*}{Time} & \multicolumn{4}{c}{PanTCR-GF2}  & \multicolumn{4}{c}{WV3}  
		\\ \cmidrule(r){5-8} \cmidrule(r){9-12}
            &(M)$\downarrow$ &(G)$\downarrow$ &(s)$\downarrow$ 
		& PSNR($\uparrow$)    & SSIM($\uparrow$)    & SAM($\downarrow$)    & ERGAS($\downarrow$)   
		& PSNR($\uparrow$)    & SSIM($\uparrow$)    & SAM($\downarrow$)    & ERGAS($\downarrow$)  \\ 
		\midrule
	TV
	& -    & -    & 24.07
	& 25.50 (-)    & 0.725 (-)    & 3.05 (-)    & 4.85 (-) 
        & 30.28 (-)    & 0.815 (-)    & 6.14 (-)    & 6.23 (-) \\
        
    LRTCFPan
    & -    & -    & 38.24
	& 23.96 (-)    & 0.633 (-)    & 3.19 (-)    & 5.74 (-)    
        & 31.71 (-)    & 0.876 (-)    & 5.46 (-)    & 5.49 (-) \\
    SRPPNN
    & 0.898    & 21.080    & \underline{0.006}
	& 27.64 (26.95)    & 0.823(0.782)     & 2.86 (3.06)     & 3.83 (4.10)    
        & 37.41 (37.35)    & 0.965 (0.963)    & 3.45 (3.50)     & 2.67 (2.48) \\
        
    BiMPan
    & 0.603    & 6.480     & 0.011
	& 27.75 (27.39)    & 0.827 (0.816)    & 2.79 (3.01)     & 3.77 (3.92)    
        & 37.22 (34.27)    & 0.963 (0.937)    & 3.56 (4.79)     & 2.59 (3.58) \\

    FusionMa.
    & 0.726    & \underline{1.360}     & 0.430
    & 27.69 (28.12)    & 0.822 (0.836)    & \underline{2.77} (2.98)     & 3.80 (3.67)    
    & 37.00 (37.37)    & 0.959 (0.963)    & 3.61 (3.53)     & 2.62 (2.51) \\
        
	DISPNet
    & 1.567    & 27.668     & 0.009
	& 27.59 (26.77)    & 0.819 (0.776)    & 2.78 (3.08)     & 3.84 (4.19)    
        & 37.59 (37.57)    & 0.964 (0.963)    & 4.38 (3.35)     & 3.40 (2.45) \\

    SSDiff
    & 1.288    & 35.180    & 4.376
    & 27.58 (27.81)    & 0.818 (0.828)    & 2.79 (3.06)     & 3.86 (3.70)    
    & 37.51 (37.62)    & 0.964 (0.967)    & 3.42 (3.49)     & 2.54 (2.41) \\

	SFINet
    & 1.770    & 11.452     & 0.015
	& 27.74 (27.36)    & 0.822 (0.810)    & 2.87 (3.01)     & 3.79 (3.89)    
        & 35.45 (35.77)    & 0.947 (0.948)    & 4.52 (4.52)     & 3.19 (3.04) \\
        
    PanMamba
    & \underline{0.478}    & 3.008   & 0.025  
        & 27.65 (27.95)    & 0.820 (0.831)    & 2.79 (2.95)     & 3.82 (3.67)    
        & 36.97 (37.82)    & 0.958 (0.964)    & 3.35 (3.40)     & 2.76 (2.44) \\
       	 								
    PreMix
    & 2.194    & 21.406     & 0.014
	& 27.67 (27.72)    & 0.822 (0.830)    & 2.78 (2.98)     & 3.82 (3.62)    
        & 37.75 (38.13)    & 0.966 (0.967)    & 3.25 (3.36)     & 2.38 (2.31) \\

    ARConv
    & 4.762    & 15.35     & 0.111
	& 27.67 (27.87)    & 0.819 (0.838)    & 2.82 (3.06)     & 3.82 (3.75)    
        & 37.30 (37.74)    & 0.962 (0.966)    & 3.34 (3.60)     & 2.52 (2.47) \\
        
    WFANet
    & 0.507    & 3.496     & 0.043
	& 27.93 (\underline{28.32})    & 0.829 (\underline{0.852})    & 2.81 (2.88)     & 3.73 (\underline{3.54})    
        & 37.88 (\underline{38.37})    & 0.966 (\underline{0.970})    & 3.18 (\underline{3.34})     & 2.34 (\underline{2.22}) \\
    Ours
    & \textbf{0.310}    & \textbf{0.901}  & \textbf{0.005}
        & \textbf{29.08}    & \textbf{0.874}    & \textbf{2.75}    & \textbf{3.23}  
        & \textbf{38.77}    & \textbf{0.972}    & \textbf{3.04}    & \textbf{2.11}  \\
		\bottomrule[1.2pt] 
	\end{tabular}}
    \caption{Quantitative results on PanTCR-GF2 and WV3 with reduced-resolution test data. Metrics are reported for two-stage processing (outside brackets) and end-to-end processing (inside brackets).
    \textbf{Note:} All two-stage methods need to go through two decloud networks first, so the \textbf{“Param"} and \textbf{“FLOPs"} columns should add \textbf{6.823M} and \textbf{8.435G}, respectively.}
	\label{table_reduced_resolution}
\end{table*}

\section{Experiment}
\subsection{Datasets and Benchmark}
We evaluate our method on real and simulated data: the real-world PanTCR-GF2 dataset and a synthetic WV3 dataset (from PanCollection) degraded by thin clouds with four morphologies and varying thicknesses.

We compare our method with cutting-edge deep learning methods, including SRPPNN~\cite{srppnn}, BiMPan~\cite{bimpan}, SFINet~\cite{sfinet}, PreMix~\cite{premix}, DISPNet~\cite{dispnet}, FusionMamba~\cite{fusionmamba}, PanMamba~\cite{panmamba}, SSDiff~\cite{ssdiff}, ARConv~\cite{arconv}, and WFANet~\cite{wfanet}, and two traditional-based methods (TV~\cite{tv} and LRTCFPan~\cite{lrtcfpan}). 

We evaluate comparison methods under two settings: (1) \textbf{Two-stage setting}, where a state-of-the-art (SOTA) thin-cloud removal model (ThiefCloud~\cite{thiefcloud}) preprocesses both panchromatic (PAN) and low-resolution multi-spectral image (LR-MSI), respectively, followed by the compared pansharpening methods. Specifically, the cloudy LR-MSI and cloudy PAN serve as the input to the decloud stage, with their corresponding clean versions (clean LR-MSI and PAN) as ground truth (GT), respectively. The output decloud PAN and LR-MSI are then used as inputs for pansharpening, supervised by the clean high-resolution MSI (HR-MSI). (2) \textbf{End-to-end setting}, where deep learning-based methods are directly applied to the cloudy images as unified pansharpening solutions, supervised by the clean HR-MSI. Note that the end-to-end setting does not apply to traditional methods. 

We evaluate pansharpening methods by assessing only their final HR-MSI outputs in reduced- and full-resolution settings. For the reduced-resolution setting, we use quantitative reference-based metrics, including PSNR, SSIM, SAM~\cite{alparone2007comparison}, and ERGAS~\cite{wald2002data}, which are commonly adopted for both pansharpening and thin-cloud removal tasks. For the full-resolution setting, we employ widely used no-reference pansharpening metrics: $D_\lambda$, $D_s$, and HQNR~\cite{alparone2008multispectral}. To further evaluate the perceptual cloud removal quality of the final HR-MSI~\cite{sun2026align}, we also report BRISQUE~\cite{mittal2012no}, and NIQE~\cite{mittal2012making}, three no-reference metrics commonly used for assessing image degradation and restoration quality.

\subsection{Implement Details}
Our algorithm was implemented in the PyTorch framework, with all experiments conducted on an NVIDIA 4090D GPU. We adopt the \(l_1\) loss function between the predicted high-resolution MSI and the corresponding GT for optimization and utilize the Adam optimizer \cite{kingma2014adam,prcv} with parameters $\beta_1=0.9$, $\beta_2=0.999$, and $\epsilon=10^{-8}$. A cosine annealing schedule is employed to decay the learning rate from $3e-3$ to $1e-6$ over 500 epochs~\cite{10516464,SUN2026112800}.

\subsection{Comparison with State-of-the-Art Methods}
Table~\ref{table_reduced_resolution} and Table~\ref{table_full_size} report quantitative results on the PanTCR-GF2 and WV3 datasets under both reduced-resolution and full-resolution settings, evaluating both a \textbf{two-stage pipeline} and an \textbf{end-to-end setting}.

From the results, we observe that deep learning-based methods with strong generalization capacity, such as WFANet and ARConv, tend to perform well in the end-to-end setting on structure-focused metrics (e.g., PSNR, SSIM). However, due to the absence of explicit cloud removal mechanisms, these methods exhibit inferior performance on quality-aware or spectral fidelity metrics (e.g., SAM, NIQE) compared with the two-stage setting, reflecting their limited ability to address cloud-induced degradation. Conversely, lightweight models like SRPPNN and BiMPan suffer a more significant performance drop in the end-to-end mode, likely due to insufficient representational capacity for modeling complex degradations~\cite{li2025pointdico}. In contrast, our proposed unified framework consistently achieves superior results across all major metrics, both in reduced- and full-resolution settings.

\begin{figure}
    \centering
    \includegraphics[width=1\linewidth]{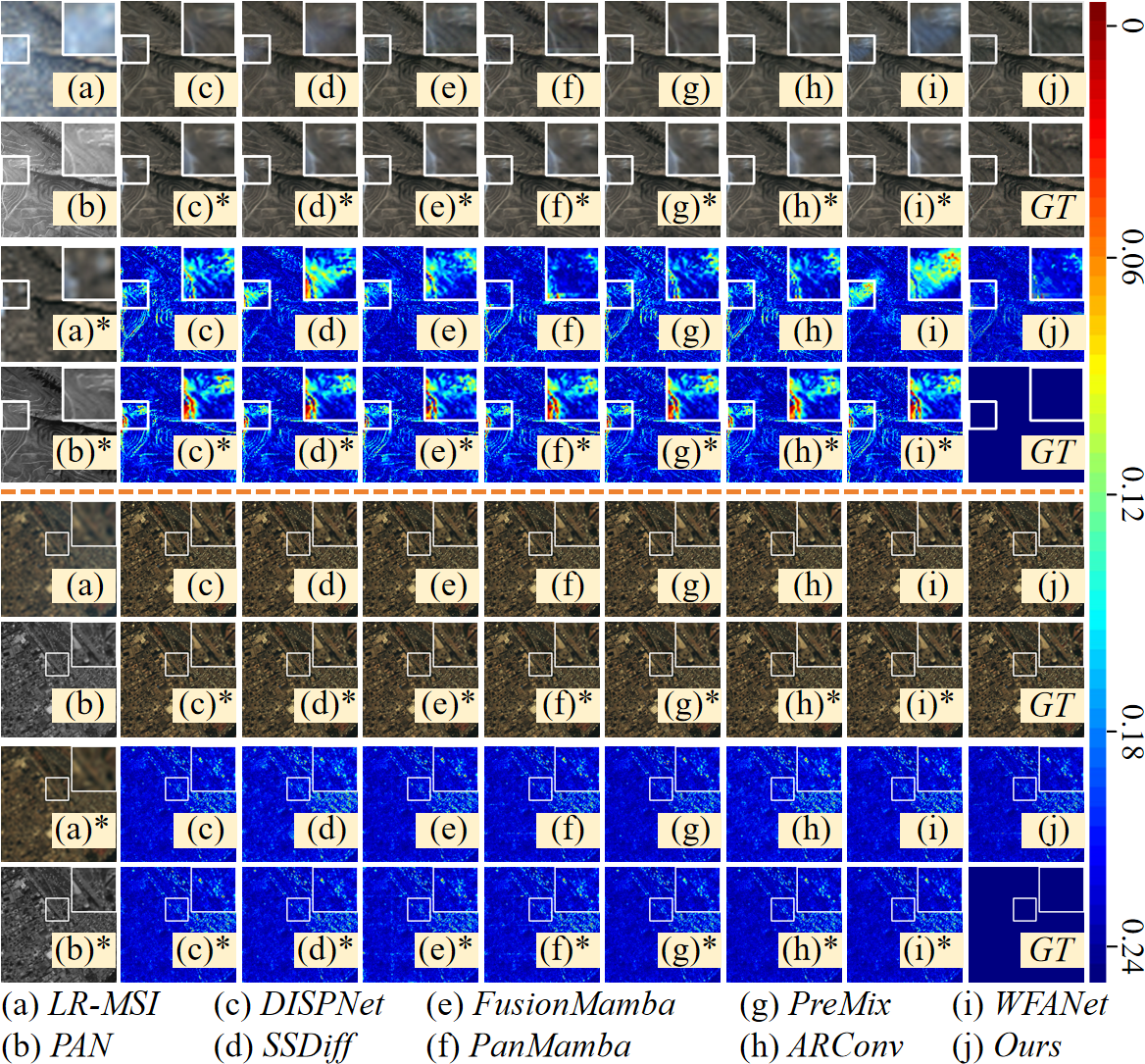}
    \caption{Qualitative results on PanTCR-GF2 and WV3 datasets with reduced-resolution, including restored images and residual error maps. Inputs (“LR-MSI", “PAN") or methods marked with “*" indicate two-stage pipelines, where cloud removal is performed prior to pansharpening.}
    \label{fig_visual_results}
\end{figure}

Furthermore, qualitative comparisons in Fig.~\ref{fig_visual_results} further support our findings. Residual maps (MSE between outputs and GT) show that our results have the darkest residuals, indicating superior fidelity over all baselines.

\begin{table*}[]
	\centering
	\resizebox{\linewidth}{!}{\begin{tabular}{ccccccccccc}
		\toprule[1.2pt]
		\multirow{2}{*}{Method} & \multicolumn{5}{c}{PanTCR-GF2}  & \multicolumn{5}{c}{WV3}  
		\\ \cmidrule(r){2-6} \cmidrule(r){7-11}
		& $D_\lambda$($\downarrow$)    & $D_s$($\downarrow$)    & HQNR($\uparrow$)    & BRISQUE($\downarrow$)     & NIQE($\downarrow$)    
        & $D_\lambda$($\downarrow$)    & $D_s$($\downarrow$)    & HQNR($\uparrow$)    & BRISQUE($\downarrow$)     & NIQE($\downarrow$)    \\ 
		\midrule
        
        SRPPNN
	& 0.033 (0.048)    & 0.178 (0.147)    & 0.795 (0.812)   & 44.05 (44.59)   & 5.13 (5.43)   
        & 0.038 (0.029)    & 0.041 (0.055)    & 0.923 (0.918)   & 34.06 (35.11)   & 4.22 (4.25)   \\

        BiMPan
	& 0.034 (0.042)    & 0.158 (0.128)    & 0.813 (0.835)   & 46.04 (45.46)   & 5.58 (5.65)   
        & 0.032 (0.055)    & 0.044 (0.060)    & 0.926 (0.889)   & 31.91 (\textbf{29.50})   & 4.26 (4.60)   \\

        FusionMa.
	& 0.039 (0.041)    & 0.177 (0.129)    & 0.791 (0.836)   & 46.32 (46.04)   & 5.32 (5.01)   
        & 0.033 (\textbf{0.027})    & 0.046 (0.055)    & 0.923 (0.919)   & 47.83 (34.85)   & 5.64 (4.19)   \\
        					
	DISPNet
        & 0.035 (0.041)    & 0.179 (0.139)    & 0.792 (0.826)   & 48.79 (44.84)   & 5.67 (4.96)   
	& 0.037 (0.032)    & 0.043 (0.068)    & 0.922 (0.903)   & 35.33 (34.42)   & 4.29 (4.42)   \\

        SSDiff
	& 0.037 (0.041)    & 0.157 (0.120)    & 0.812 (0.844)   & 44.98 (44.49)   & 5.50 (5.64)   
        & 0.032 (0.039)    & 0.045 (0.045)    & 0.925 (0.918)   & 34.33 (36.08)   & 4.53 (4.47)   \\
        					
	SFINet
	& \textbf{0.031} (0.045)    & 0.179 (0.125)    & 0.796 (0.836)   & 47.00 (47.20)    & 5.74 (5.21)   
        & 0.033 (0.048)    & 0.042 (0.044)    & 0.926 (0.910)   & 33.98 (34.51)    & 4.32 (4.34)   \\

        PanMamba
	& 0.033 (0.039)    & 0.165 (0.144)    & 0.807 (0.823)   & \underline{44.35} (44.73)   & 4.96 (4.93)  
        & 0.036 (0.029)    & 0.045 (0.057)    & 0.921 (0.916)   & 32.34 (32.69)   & \textbf{4.07} (4.19)  \\
				
        PreMix
	& 0.035 (0.046)    & 0.175 (0.142)    & 0.796 (0.819)   & 44.88 (45.54)   & 5.51 (5.48)   
        & 0.029 (0.030)    & 0.041 (0.047)    & \underline{0.931} (0.924)   & 34.51 (37.76)   & 4.22 (4.51)   \\

        ARConv
	& 0.036 (0.043)    & 0.188 (0.127)    & 0.782 (0.837)   & 46.50 (46.06)   & \underline{4.92} (5.10)   
        & 0.031 (0.028)    & \underline{0.041} (0.046)    & 0.929 (0.927)   & 32.95 (34.84)   & 4.28 (4.34)   \\

        WFANet
	& 0.036 (0.040)    & 0.167 (\underline{0.118})    & 0.803 (\underline{0.847})   & 45.26 (45.30)   & 5.16 (5.36)   
        & 0.029 (0.030)    & 0.044 (0.059)    & 0.928 (0.913)   & 32.14 (32.61)   & 4.32 (4.41)   \\

        Ours
        & \underline{0.033}    & \textbf{0.117}    & \textbf{0.854}   & \textbf{43.91}    & \textbf{4.84}     
        & \underline{0.028}    & \textbf{0.040}    & \textbf{0.933}   & \underline{31.04}    & \underline{4.13}    \\

		\bottomrule[1.2pt] 
	\end{tabular}}
    \caption{Quantitative results on the PanTCR-GF2 and WV3 with full-resolution test data. Metrics are reported for two-stage processing (outside brackets) and end-to-end processing (inside brackets).}
	\label{table_full_size}
\end{table*}

\subsection{Ablation Experiments}
To comprehensively evaluate the effectiveness of each component in our framework, we conduct ablation studies from four key perspectives: (1) degradation modeling, (2) frequency-domain modeling, (3) spectral enhancement, and (4) learning strategy. As summarized in Table~\ref{table_ablation}, experiments are performed on the PanTCR-GF2 dataset using both reduced- and full-resolution test data, with metrics including PSNR, SSIM, and SAM (reduced-resolution), and HQNR and BRISQUE (full-resolution).

1) \textbf{Degradation Modeling.} We first remove the two Degradation Aware Modules (DAMs) equipped in the frequency decoupled restoration (FDR) block, and directly fuse amplitude (\(\mathcal{A}_\mathcal{I}\),\(\mathcal{A}_\mathcal{F}\)) and phase (\(\mathcal{P}_\mathcal{X}\),\(\mathcal{P}_\mathcal{F}\)) components using concatenation (denoted as ‘w/o Deg. Aware’). This results in a consistent degradation across all metrics, indicating that explicit modeling of degradation priors is essential for robust cloud removal and spatial detail preservation. To further investigate the role of degradation-robust prompts, we conduct three additional variants: removing the PAN or NIR individually (‘w/o Pan’), only the NIR amplitude guidance (‘w/o NIR’), and both (‘w/o Pan\&NIR’). All variants lead to performance drops, especially in SAM and ERGAS, validating the complementary nature of NIR’s cloud robustness and PAN’s spatial richness in guiding the restoration.

2) \textbf{Frequency-Domain Modeling.} We first assess the effectiveness of the high-pass filter \(\mathcal{H}(\cdot)\), which is designed to extract high-frequency details from PAN and suppress thin-cloud interference (‘w/o \(\mathcal{H}(\cdot)\)'), proving critical for preserving structural fidelity under cloud-contaminated conditions. We then evaluate the effectiveness of frequency-specific branches by individually disabling the phase prompt-guided branch (‘w/o Pha. Branch’) and the amplitude prompt-guided branch (‘w/o Amp. branch’), while retaining the cross-modulation from the interactive inter-frequency consistency (IFC) module. Removing the phase prompt-guided Branch leads to a noticeable loss in structural fidelity (\(\downarrow\)PSNR, \(\downarrow\)SSIM), whereas removing the amplitude prompt-guided branch affects spectral consistency and cloud suppression (\(\uparrow\)SAM, \(\uparrow\)BRISQUE). This confirms the necessity of decoupled processing for targeted restoration.

\begin{table}
\centering
\resizebox{\linewidth}{!}{\begin{tabular}{c|c|ccc|cc} 
\toprule[1.2pt]
                    & Configurations   & PSNR  & SSIM  & SAM   & HQNR  & BRISQUE  \\ 
\midrule
\multirow{4}{*}{1}  & w/o Deg. Aware      & 27.96     & 0.843     & 2.95      & 0.826     & 45.32  \\
                    & w/o Pan             & 28.32     & 0.856     & 2.86      & 0.838     & 44.43  \\
                    & w/o NIR             & 28.46     & 0.862     & 2.93      & 0.842     & 44.96  \\ 
                    & w/o Pan\&NIR        & 28.03     & 0.848     & 2.96      & 0.823     & 45.65  \\ 
\midrule              
\multirow{7}{*}{2}  
              & w/o $\mathcal{H}(\cdot)$  & 28.33     & 0.853     & 2.79      & 0.844     & 44.07  \\
                    & w/o Pha. Branch     & 27.94     & 0.838     & 2.81      & 0.828     & 44.77  \\
                    & w/o Amp. Branch     & 28.38     & 0.852     & 2.92      & 0.839     & 45.48  \\
        & FDR$\rightarrow$ Spa-Atten.     & 27.94     & 0.840     & 2.96      & 0.825     & 46.06  \\
                    & w/o FDR             & 27.86     & 0.823     & 3.07      & 0.826     & 46.33  \\
                    & w/o MAFG            & 28.45     & 0.859     & 2.88      & 0.848     & 45.19  \\ 
                    & w/o IFC             & 28.03     & 0.848     & 2.90      & 0.829     & 45.43  \\
        & IFC$\rightarrow$Cha-Atten.      & 28.56     & 0.866     & 2.84      & 0.845     & 43.96  \\
        & Parallel$\rightarrow$ Series    & 28.09     & 0.846     & 3.04      & 0.838     & 45.06  \\
\midrule
\multirow{2}{*}{3}  & w/o SE              & 27.96     & 0.841     & 3.06      & 0.829     & 44.89  \\
         & SE $\rightarrow$Cha-Atten.     & 28.22     & 0.853     & 2.88      & 0.839     & 44.43  \\
\midrule
\multirow{1}{*}{4}  
& Unified$\rightarrow$Two-stage           & 28.36     & 0.853     & 2.78      & 0.842     & 44.03  \\
\midrule                                  
                    & Full model
  & \textbf{29.08} & \textbf{0.874} & \textbf{2.75}  & \textbf{0.854}    &\textbf{43.91}  \\
\bottomrule[1.2pt]
\end{tabular}}
\caption{Ablation study results with the best results in \textbf{bold}.}
\vspace{-0.5cm}
\label{table_ablation}
\end{table}

Further, we explore two structural variants: (1) replacing the entire FDR block with a spatial attention module (‘FDR\(\rightarrow\)Spa-Atten.’), and (2) removing the FDR block entirely (‘w/o FDR’). Both result in sharp performance degradation across all metrics, highlighting the superiority of frequency-based modeling over purely spatial baselines.

We also ablate the modality-adaptive frequency gating (MAFG) unit (‘w/o MAFG’), which leads to reduced fusion quality due to inadequate suppression of modality-induced spectral bias. Additionally, we validate the necessity of the IFC module: replacing it with vanilla channel attention (‘IFC\(\rightarrow\)Cha-Atten.’) or removing it completely (‘w/o IFC’) results in degraded performance and misaligned frequency fusion. We also test a serial configuration of the two parallel branches (‘Parallel\(\rightarrow\)Series’), confirming that our parallel design enables better decoupling and efficiency.

3) \textbf{Spectral Enhancement.} To assess the role of the spectral enhancement (SE) module, we remove it (‘w/o SE’) and replace it with channel attention (‘SE\(\rightarrow\)Cha-Atten.’), while retaining the FDR block. Both modifications lead to notable performance degradation, particularly on SAM and ERGAS, suggesting that although frequency-based decoupling effectively restores structure and handles degradation, it may fail to recover fine-grained spectral fidelity. The SE module, built on spectral-wise transformer (SWT), complements frequency restoration by introducing localized spectral corrections, thus improving spectral coherence.

4) \textbf{Unified Framework.} Finally, we convert our end-to-end unified framework with a cascaded two-stage variant (‘Unified→Two-stage’). Despite increased model complexity, the performance declines, especially on cloud-relevant and perceptual metrics, emphasizing the importance of unified joint modeling in enhancing overall performance.

\section{Conclusion}
In this paper, we present a unified and lightweight framework that simultaneously addresses thin-cloud removal and pansharpening, effectively tackling the dual degradations caused by cloud contamination and spatial resolution limitations. By leveraging frequency-domain priors from PAN and NIR modalities, our method explicitly decouples amplitude and phase components and performs targeted restoration, enabling accurate cloud suppression and structural detail enhancement within a single end-to-end network. In contrast to conventional two-stage pipelines, our approach offers a computationally efficient solution while achieving superior reconstruction quality. To further support this task, we construct \textbf{PanTCR-GF2}, the first real-world dataset tailored for thin-cloud pansharpening, which fills a critical gap in remote sensing benchmarks where cloud-related degradations have long been underexplored. Extensive experiments demonstrate the superiority of our unified approach over existing algorithms in both accuracy and efficiency.

\bibliography{aaai2026}


\end{document}